\documentclass[conference,a4paper]{IEEEtran}
\IEEEoverridecommandlockouts

\ifCLASSINFOpdf
\else
\fi

\usepackage[dvips]{graphicx}
\usepackage{indent}
\usepackage{amsmath,amssymb}
\usepackage{paralist}
\usepackage{eclbkbox}
\usepackage{subfigure}
\usepackage{url}
\usepackage{algorithm}
\usepackage{algorithmic}

\makeatletter
\def\tbcaption{\def\@captype{table}\caption}
\def\figcaption{\def\@captype{figure}\caption}

\newcommand{\bvec}[1]{\mbox{\boldmath $#1$}}

\begin{document}
\title{Adaptive Learning Method of Recurrent Temporal Deep Belief Network to Analyze Time Series Data
\thanks{\copyright 2017 IEEE. Personal use of this material is permitted. Permission from IEEE must be obtained for all other uses, in any current or future media, including reprinting/republishing this material for advertising or promotional purposes, creating new collective works, for resale or redistribution to servers or lists, or reuse of any copyrighted component of this work in other works.}
}

\author{
\IEEEauthorblockN{Takumi Ichimura}
\IEEEauthorblockA{Faculty of Management and Information Systems,\\
Prefectural University of Hiroshima\\
1-1-71, Ujina-Higashi, Minami-ku,\\
Hiroshima, 734-8559, Japan\\
Email: ichimura@pu-hiroshima.ac.jp}
\and
\IEEEauthorblockN{Shin Kamada}
\IEEEauthorblockA{Graduate School of Information Sciences, \\
Hiroshima City University\\
3-4-1, Ozuka-Higashi, Asa-Minami-ku,\\
Hiroshima, 731-3194, Japan\\
Email: da65002@e.hiroshima-cu.ac.jp}
}

\maketitle

\begin{abstract}
Deep Learning has the hierarchical network architecture to represent the complicated features of input patterns. Such architecture is well known to represent higher learning capability compared with some conventional models if the best set of parameters in the optimal network structure is found. We have been developing the adaptive learning method that can discover the optimal network structure in Deep Belief Network (DBN). The learning method can construct the network structure with the optimal number of hidden neurons in each Restricted Boltzmann Machine and with the optimal number of layers in the DBN during learning phase. The network structure of the learning method can be self-organized according to given input patterns of big data set. In this paper, we embed the adaptive learning method into the recurrent temporal RBM and the self-generated layer into DBN. In order to verify the effectiveness of our proposed method, the experimental results are higher classification capability than the conventional methods in this paper.
\end{abstract}

\IEEEpeerreviewmaketitle

\section{Introduction}
\label{sec:Introduction}
Recently, Deep Learning attracts a lot of attention in methodology research of artificial intelligence such as machine learning \cite{Bengio09}. The learning architecture has an advantage of not only multi-layered network structure but also actualization of pre-training. The pre-training realizes that the architecture of Deep Learning accumulates prior knowledge of the features for input patterns. The convolutional neural network (CNN) is well known to be feed forward neural network and the layers of CNN have neurons arranged in 3 dimensions. However, the building a new CNN takes much costs to find an optimal structure for the given input patterns and the CNN with the higher detection capability for the unknown patterns can not be constructed because the optimal set of many parameters is not found instantly.

On the other hand, Restricted Boltzmann Machine (RBM) \cite{Hinton12} and Deep Belief Network (DBN) \cite{Hinton06} are focused as one of popular method of Deep Learning for unsupervised learning. RBM has the capability of representing a probability distribution of input data set, and it can represent an energy-based statistical model. Moreover, the Contrastive Divergence (CD) learning procedure  can be often used as one of the learning methods of RBM \cite{Hinton02,Tileman08}. CD method is a faster algorithm of Gibbs sampling based on Markov chain Monte Carlo method. However, the problem of RBMs is also the definition of an optimal initial network structure including the optimal number of hidden neurons according to the features of input patterns.

DBN has a deep architecture that can represent multiple features of input patterns hierarchically with the pre-trained RBM. In other word, each layer of DBN employs RBM learning method to implement the pre-training and then DBN consists of the hierarchical two or more trained RBMs. RBM at lower layer can represent the abstract character, while one at higher layer shows the concrete object. If the energy function at each layer are good performance, the whole network can achieve the higher capability than the traditional neural network.

We proposed the adaptive learning method of RBM that can construct a RBM with an optimal number of hidden neurons according to the training situation by applying the neuron generation and annihilation algorithm \cite{Kamada16_ICONIP16,Kamada16_SMC2016}. Moreover, we developed the hierarchical adaptive learning method of DBN that can determine the optimal number of hidden layers according to the given data set \cite{Kamada16_TENCON16}. The layer generation of DBN is natural extension to develop the adaptive learning method with self-organizing mechanism.

Recently, the analysis of high dimensional input patterns such as images in video, sounds in music is required as a new data set of Big data. Such data are time series one which consist of two or more sequences of discrete time data. A general time series data are frequently plotted by using line charts such as signal processing, weather forecasting and mathematical finances. However, high dimensional data at each time step such as video and music have the multi modal conditional distributions. The active researchers pursue the developing model of such sequences to predict the conditional distribution at the next time by using the previous time step data.

In recurrent neural networks, there are two variants of RNNs for modeling slot sequences: the Elman-type RNN \cite{Elman90} and the Jordan-type RNN \cite{Jordan86}. In the deeplearning field, tempral RBM and RNN-RBN are the current main stream researches. Recurrent Temporal Restricted Boltzmann Machine (RTRBM) \cite{Sutskever08} is a probabilistic model for high dimensional sequences and a directed graphical model consisting of a sequence of RBMs. The learning consists of a conditional RBM at each time step. Despite its simplicity, this model successfully accounts for several interesting sequences. Recurrent Neural Networks RBM (RNN-RBM) \cite{Lewandowski12} is a similar model of RTRBM and realizes the specification of recurrent neural networks (RNNs), which can incorporate an internal memory that can summarize the entire sequence history.

We propose the adaptive learning method in RNN-RBM to embed the neuron generation / annihilation. Moreover, we employ the hierarchical adaptive learning method of DBN that can determine the optimal number of hidden layers according to the given data set. The learning method can perform more accuracy to input patterns even if the pattern has noise. Our proposed adaptive learning method of RNN-DBN has a deep architecture that can represent multiple features of input patterns hierarchically with the pre-trained adaptive learning method of RNN-RBM. In order to verify the effectiveness of our proposed method, we investigate the prediction capability for some benchmark data sets. As a result, the adaptive learning method of RNN-DBN shows the higher score than the traditional methods.

\section{Adaptive Learning Method of RBM}
\label{sec:AdaptiveRBM}
\subsection{An Overview of RBM}
\begin{figure}[tbp]
\begin{center}
\includegraphics[scale=0.7]{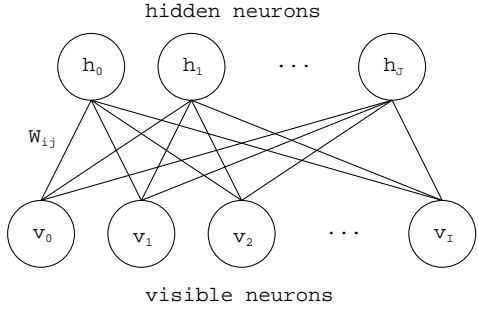}
\caption{An overview of RBM}
\label{fig:rbm}
\end{center}
\end{figure}

Fig.~\ref{fig:rbm} shows a structure of RBM which has 2 kinds of layers. The lower layer is a visible layer for input data. The upper layer is a hidden layer for representing the features to train the input data space. The neuron in both layers is a binary neuron and their connections between neurons in same layer are  removed. RBM learning method trains some parameters including weights and bias  till the energy function reaches the expected value. The trained RBM can represent the model to be fit for the probability distribution of input data space.

We explain the general RBM model mathematically in the section. Let $v_{i} (0 \leq i \leq I)$ and $h_{j} (0 \leq j \leq J)$ be a binary variable for each neuron. $I$ and $J$ are the number of visible and hidden neurons, respectively. Eq.(\ref{eq:energy}) is the energy function $E(\bvec{v}, \bvec{h})$ for visible vector $\bvec{v} \in \{ 0, 1 \}^{I}$ and hidden vector $\bvec{h} \in \{ 0, 1 \}^{J}$. Eq.(\ref{eq:prob}) is the probability distribution for \bvec{v} and \bvec{h}.
\begin{equation}
E(\bvec{v}, \bvec{h}) = - \sum_{i} b_i v_i - \sum_j c_j h_j - \sum_{i} \sum_{j} v_i W_{ij} h_j ,
\label{eq:energy}
\end{equation}
\begin{equation}
p(\bvec{v}, \bvec{h})=\frac{1}{Z} \exp(-E(\bvec{v}, \bvec{h})), \;  Z = \sum_{\bvec{v}} \sum_{\bvec{h}} \exp(-E(\bvec{v}, \bvec{h})) ,
\label{eq:prob}
\end{equation}
where $b_i$ and $c_j$ are the bias parameters for $v_i$ and $h_j$, respectively. $W_{ij}$ is the weight between $v_i$ and $h_j$. $Z$ is the partition function by summing over all possible pairs of visible and hidden vectors. The RBM can learn the biases and weights $\bvec{\theta}=\{\bvec{b}, \bvec{c}, \bvec{W} \}$ according to the distribution of input data by Contrastive Divergence (CD) \cite{Hinton02}. CD method is a faster algorithm of Gibbs sampling based on Markov chain Monte Carlo methods and can realize a good performance in smaller number of sampling steps.

The RBM learning method by CD method should consider the convexity and continuous conditions for an objective function. Carlson et al. et al. derived that the RBM learning by CD method will be converged if the variance for 3 kinds of parameters $\bvec{\theta}=\{\bvec{b}, \bvec{c}, \bvec{W} \}$ falls into a certain range during the training under the Lipschitz continuous (please see Eq.(18)-(20) in \cite{Carlson15} for details).

Under the convexity and continuous conditions, we observed the oscillation of gradients of three biases $\bvec{b}$, $\bvec{c},$ and $\bvec{W}$ during the learning phase. From some results on benchmark datasets, the change for the two kinds of parameters $\bvec{c}$ and $\bvec{W}$ is very large and then we found the regularity to avoid the repercussion of uniformity of impact data \cite{Kamada16_SMC2016}.

\subsection{Neuron Generation and Annihilation Algorithm}
\label{subsec:neurongeneration}
We have proposed the adaptive learning method of RBM that the optimal number of hidden neurons can be self-organized according to the features of a given input data set in learning phase. The neuron generation and annihilation algorithm of RBM can measure the criterion of network stability with the fluctuation of weight vector \cite{Kamada16_ICONIP16,Kamada16_SMC2016}.

\begin{figure}[tbp]
\begin{center}
\subfigure[Neuron Generation]{\includegraphics[scale=0.5]{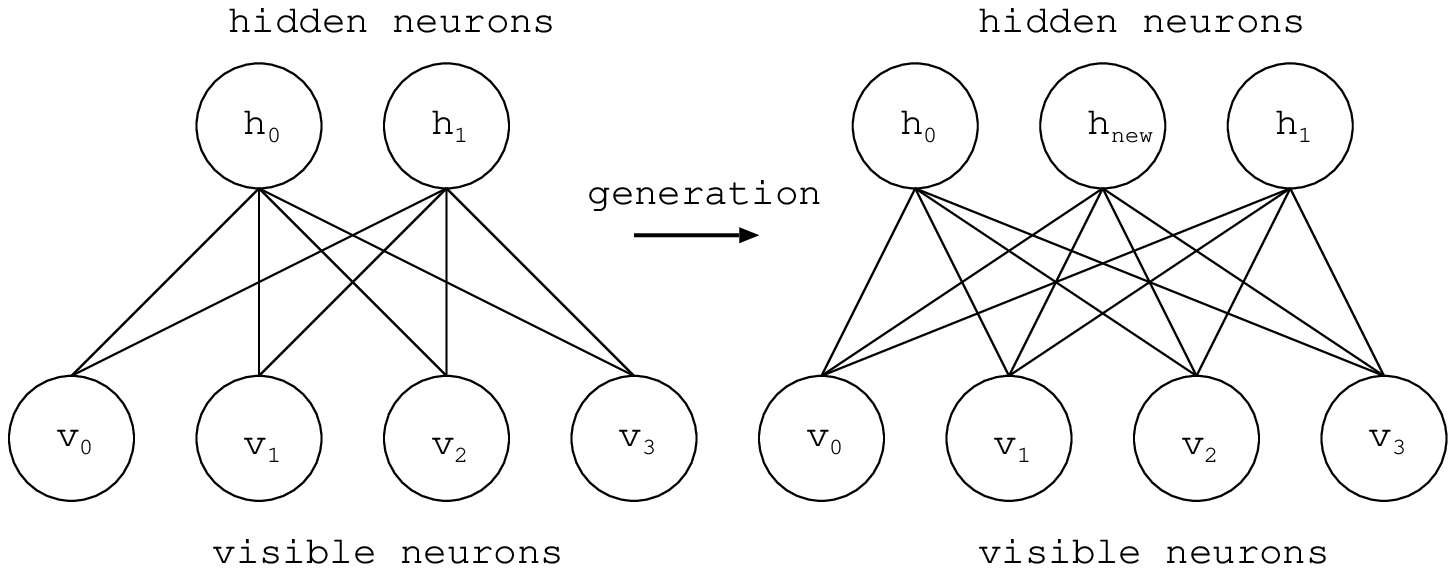}\label{fig:neuron_generation}}
\subfigure[Neuron Annihilation]{\includegraphics[scale=0.5]{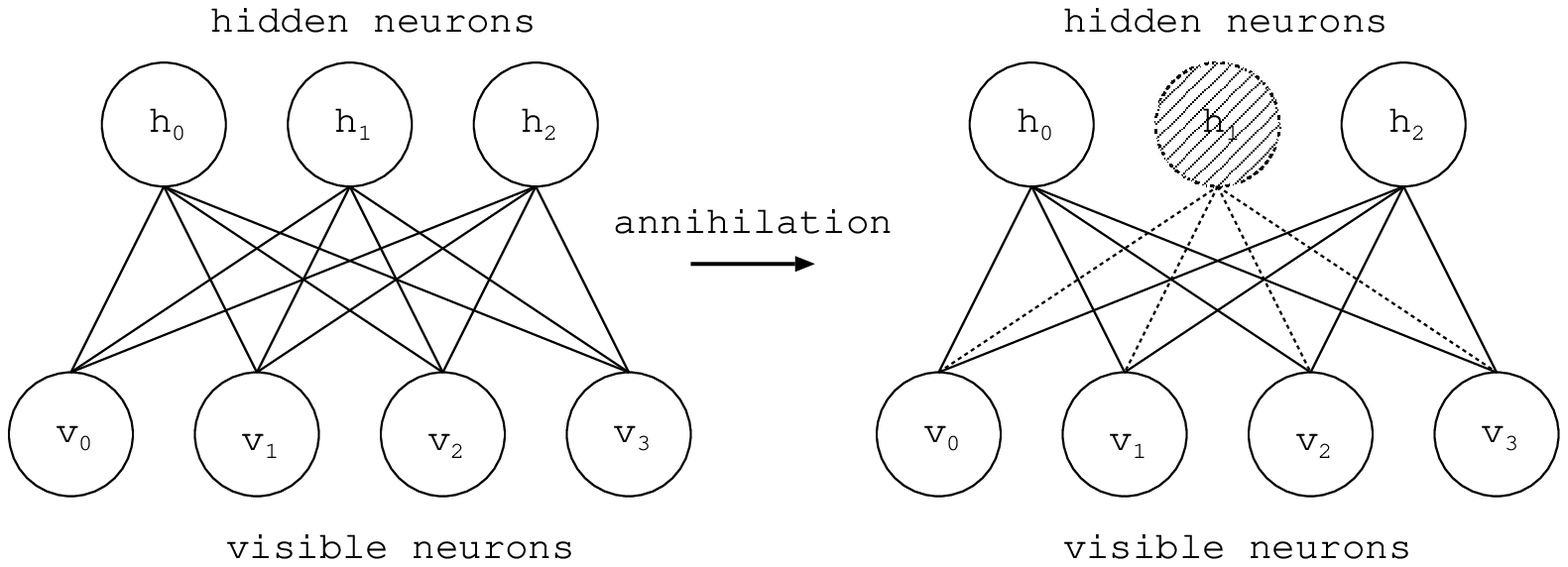}\label{fig:neuron_annihilation}}
\caption{The neuron generation/annihilation algorithm}
\label{fig:adaptive_rbm}
\end{center}
\end{figure}

The basic concept of neuron generation and annihilation algorithm is as follows. A new neuron is generated in case of the lack of classification capability of hidden neurons. The weight will fluctuate greatly even after the training process, since some hidden neurons may not represent an ambiguous pattern. In such a case, the corresponding neuron is split into 2 neurons to represent the complex pattern by inheriting the attributes of the parent hidden neuron. Then we monitor the value of left term of Eq.(\ref{eq:neuron_generation}) to monitor the condition of neuron generation.

\begin{equation}
(\alpha_{c} \cdot dc_j) \cdot (\alpha_{W} \cdot dW_{ij} )> \theta_{G}
\label{eq:neuron_generation}
\end{equation}
The equation is the inner product of variance of monitoring both of 2 parameters $\bvec{c}$ and $\bvec{W}$. $dc_j$ and $dW_{ij}$ in Eq.(\ref{eq:neuron_generation}) are the gradient vectors of the hidden neuron $j$ and the weight vector between the neuron $i$ and $j$, respectively. $\alpha_{c}$ and $\alpha_{W}$ are the constant values for the adjustment of the range of each parameter. $\theta_{G}$ is an appropriate threshold value. Fig.~\ref{fig:neuron_generation} shows that a new hidden neuron generates by insertion it into the neighborhood of the parent neuron.

After the neuron generation process, if the network with the inactivated neuron that does not work to the classification capability, the corresponding neuron should be removed in terms of the reduction of calculation. If Eq.(\ref{eq:neuron_annihilation}) is satisfied in learning phase, the corresponding neuron will be annihilated as shown in Fig.~\ref{fig:neuron_annihilation}. 

\begin{equation}
\frac{1}{N}\sum_{n=1}^{N} p(h_j = 1 | \bvec{v}_n) < \theta_{A},
\label{eq:neuron_annihilation}
\end{equation}
where $\bvec{v}_{n}=\{ \bvec{v}_{1},\bvec{v}_{2},\cdots,\bvec{v}_{N}\}$ is a given input data, $N$ is the number of samples of input data. $p(h_j = 1 | \bvec{v}_n)$ means a conditional probability of $h_j \in \{ 0, 1 \}$ under a given $\bvec{v}_n$. $\theta_{A}$ is an appropriate threshold value. The proposed method by the neuron generation and annihilation algorithm showed the good performance for some benchmark tests \cite{Kamada16_SMC2016}.

\subsection{A Structural Learning Method with Forgetting}
\label{sec:structurallearning}
Although the optimal number of hidden neurons is determined by the neuron generation and annihilation algorithms, we may meet another difficulty that the trained network is still a black box, and then we cannot extract some explicit knowledge from the trained network. In order to solve the difficulty, we developed the Structural Learning Method with Forgetting (SLF) to discover the regularities of the network structure in RBM \cite{Kamada16_ICONIP16}. The basic concept is the structural learning with forgetting method \cite{Ishikawa96} where three kinds of penalty terms are added into an original objective function $J$ as shown in Eq.(\ref{eq:forgetting1}) - Eq.(\ref{eq:forgetting3}). Each equation is for `learning with forgetting,' `hidden units clarification,' and `learning with selective forgetting', respectively.

\begin{equation}
  \label{eq:forgetting1}
  J_{f} = J + \epsilon_{1} \| \bvec{W} \|,
\end{equation}
\begin{equation}
  \label{eq:forgetting2}
  J_{h} = J + \epsilon_{2} \sum_{i} \min \{ 1 - h_i, h_i\},
\end{equation}
\vspace{-2mm}
\begin{eqnarray}
  \label{eq:forgetting3}
  J_{s} = J - \epsilon_{3} \| \bvec{W}^{'} \|, \ 
  W^{'}_{ij} = \left\{
  \begin{array}{ll}
    W_{ij}, & if \ |W_{ij}| < \theta \\
    0, & otherwise
  \end{array}
  \right.,
\end{eqnarray}
where the range of $\epsilon_{1}$, $\epsilon_{2}$ and $\epsilon_{3}$ are smaller than the predetermined small value, which are negligible small value in the whole weight vector space. After the optimal number of hidden neurons is determined by neuron generation and annihilation algorithm, both Eq.(\ref{eq:forgetting1}) and Eq.(\ref{eq:forgetting2}) should be applied simultaneously during a certain learning period. Alternatively, Eq.(\ref{eq:forgetting3}) is used instead of Eq.(\ref{eq:forgetting1}) at final learning period so as to be the large objective function.

Distillation method of Deep learning has been proposed by Hinton \cite{Hinton15}. The method can compress the knowledge in an ensemble large model of Deep Learning into a single smaller model. The ensemble implements a function from input patterns to output patterns. Instead of the models in the ensemble model of Deep Learning, they consider a way that the model is parameterized and they focus the obtained function of network. The forgetting method in this paper realizes a basic idea of distillation method in each RBM learning before the construction of DBN.

\subsection{An Adaptive Learning Method of DBN}
\label{sec:DBN}
\begin{figure}[tbp]
\begin{center}
\includegraphics[scale=0.6]{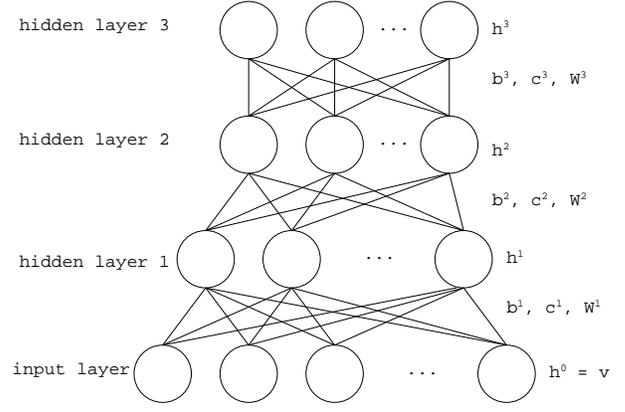}
\caption{An overview of DBN}
\label{fig:dbn}
\end{center}
\end{figure}

Fig.~\ref{fig:dbn} shows the model of Deep Belief Network (DBN) \cite{Hinton06}. The model has the hierarchical network structure where each layer is pre-trained by RBM. The hierarchical DBN network structure becomes to represent higher and multiple level features of input patterns by building up the pre-trained RBM.

In this paper, we propose the adaptive learning method of DBN that can determine the optimal number of hidden layers. The adaptive learning method of RBM is worked  by the neuron generation and annihilation algorithm described in section \ref{subsec:neurongeneration}. In general, data representation of DBN performs the specified features from abstract to concrete at each layer in the direction to output layer. That is, the lower layer has the power of non figurative representation, and the higher layer constructs the object to figure out an image of input patterns. Adaptive DBN can automatically adjust self-organization of structured data representation.

In the learning process of adaptive DBN, we observe the total WD for only the variance of both $\bvec{c}$ and $\bvec{W}$ and energy function. If the energy function is still large and the overall WD is larger than the given threshold value,
a new RBM is required to express the concrete pattern for the given input data. In other words, the large energy function and the large WD stays in the condition that the DBN has lack data representation capability to figure out an image of input patterns. Eq.(\ref{eq:layer_generation1}) and Eq.(\ref{eq:layer_generation2}) are the condition of layer generation with the total WD and the energy function.
\begin{equation}
\sum_{l=1}^{k} (\alpha_{WD} \cdot WD^{l}) > \theta_{L1},
\label{eq:layer_generation1}
\end{equation}
\vspace{-2mm}
\begin{equation}
\sum_{l=1}^{k} (\alpha_{E} \cdot E^{l}) > \theta_{L2},
\label{eq:layer_generation2}
\end{equation}
where $WD^{l}$ is the total variance of parameters $\bvec{c}$ and $\bvec{W}$ in $l$-th RBM. $E^{l}$ is the total energy function in $l$-th RBM. $k$ is the top RBM in the current DBN structure. $\alpha_{WD}$ and $\alpha_{E}$ are the constant values for the adjustment of deviant range of $WD^{l}$ and $E^{l}$. $\theta_{L1}$ and $\theta_{L2}$ are the pre-determined threshold values. If Eq.(\ref{eq:layer_generation1}) and Eq.(\ref{eq:layer_generation2}) at layer $k$ are satisfied simultaneously during learning, a new RBM $k+1$ will be generated after the learning at layer $k$. The initial values of $\bvec{b}$, $\bvec{c}$ and $\bvec{W}$ at the generated layer $k+1$ are given to be inherited from the parent(lower) RBM.

As far as the image benchmark data set CIFAR-10 and CIFAR-100 \cite{CIFAR10} as shown Table \ref{tab:result-correct-ratio-cifar10} and Table \ref{tab:result-correct-ratio-cifar100}, the classification rate for the training data set and the test dataset showed the highest capability \cite{CIFAR_record, Kamada16_SMC2016,Kamada16_TENCON16}. The determined data sets for the training and test in benchmark test are given to verify the effectiveness of the propoed learning method in the competitive evaluation.

Because a Deep Neural Network can bring back the ever seen data into an existence, the generalization power is higher than the traditional neural network. Each RBM classify the detailed patterns to represent a little difference in given data. However, the reason of the misclassification of recurrent data does not depend on the generalization power of DBN. In fact, some sequential data have the same input pattern but different output pattern in both the training data and test data.

Moreover, RBM has a restricted hidden layer which each hidden neuron does not connect each other to train the subset of input signal patterns independently. In other word, there is no intersection of subsets in hidden layer. If the input pattern is complex, RBM needs more hidden neurons where a neuron works as nonlinear signal separator for the specified pattern. Therefore, the trained RBM can incarnate the detailed data representation to prevent over fitting situation.

\begin{table}[h]
\caption{Classification Accuracy on CIFAR-10}
\vspace{-5mm}
\label{tab:result-correct-ratio-cifar10}
\begin{center}
\begin{tabular}{l|r|r}
\hline \hline
& \multicolumn{1}{c|}{Training} & \multicolumn{1}{c}{Test}  \\ \hline\hline
Traditional RBM \cite{Dieleman12} &     -     & 63.0\%  \\ \hline
Traditional DBN \cite{Krizhevsky10}      &     -     & 78.9\%  \\ \hline 
CNN \cite{Clevert15}      &     -     & 96.53\%  \\ \hline \hline
Adaptive RBM                 &   99.9\%  & 81.2\%   \\ \hline
Adaptive RBM with Forgetting  &   99.9\%  & 85.8\%   \\ \hline
Adaptive DBN with Forgetting &  100.0\%  & 92.4\%   \\
(No. layer = 5, $\theta_G = 0.05$) & & \\\hline
Adaptive DBN with Forgetting&  100.0\%  & {\bf 97.1\%}   \\
(No. layer = 5, $\theta_G = 0.01$ ) & & \\\hline
\hline 
\end{tabular}
\end{center}

\vspace{-5mm}
\end{table}
\begin{table}[h]
\caption{Classification Accuracy on CIFAR-100}
\vspace{-5mm}
\label{tab:result-correct-ratio-cifar100}
\begin{center}
\begin{tabular}{l|r|r}
\hline \hline
& \multicolumn{1}{c|}{Training} & \multicolumn{1}{c}{Test}  \\ \hline\hline
CNN \cite{Clevert15}     &     -       & 75.7\%  \\ \hline \hline
Adaptive DBN with Forgetting &  100.0\%  & 78.2\%   \\
(No. layer = 5, $\theta_G = 0.05$) & & \\\hline
Adaptive DBN with Forgetting&  100.0\%  & {\bf 81.3\%}   \\
(No. layer = 5, $\theta_G = 0.01$ ) & & \\\hline
\hline 
\end{tabular}
\end{center}
\end{table}

\section{Adaptive Learning Method of RNN-DBN}
As mentioned in section \ref{sec:Introduction}, the active researcher pursues the developing model of such sequences to predict the conditional distribution at the next time by using the previous time step data.

\begin{figure}[]
\begin{center}
\includegraphics[scale=0.6]{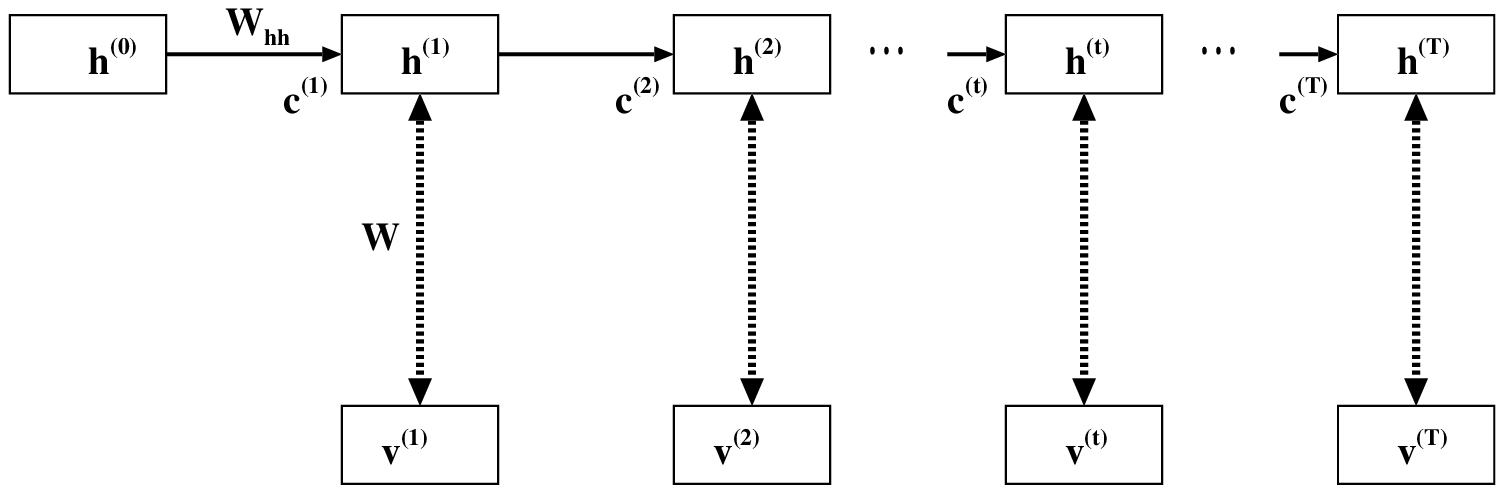}
\caption{Temporal RBM\cite{Sutskever08}}
\label{fig:trbm}
\end{center}
\end{figure}

\begin{figure}[]
\begin{center}
\includegraphics[scale=0.6]{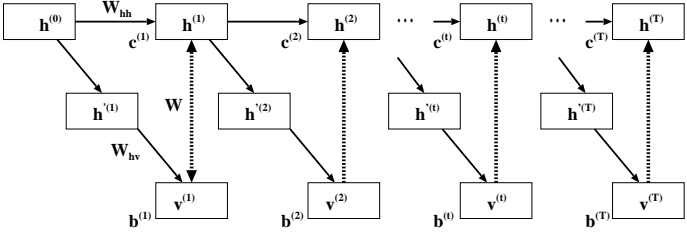}
\caption{Recurrent Temporal RBM\cite{Sutskever08}}
\label{fig:rtrbm}
\end{center}
\end{figure}

Fig.~\ref{fig:trbm} and Fig.~\ref{fig:rtrbm} show the overview of Temporal RBM and RTRBM, respectively. RTRBM \cite{Sutskever08} is a similar model of Temporal RBM. RTRBM is a probabilistic model for high dimensional sequences and forms a directed graphical model consisting of a sequence of RBMs. The learning method consists of a conditional RBM at each time step. Despite its simplicity, this model successfully accounts for several interesting sequences.

\begin{figure}[]
\begin{center}
\includegraphics[scale=0.6]{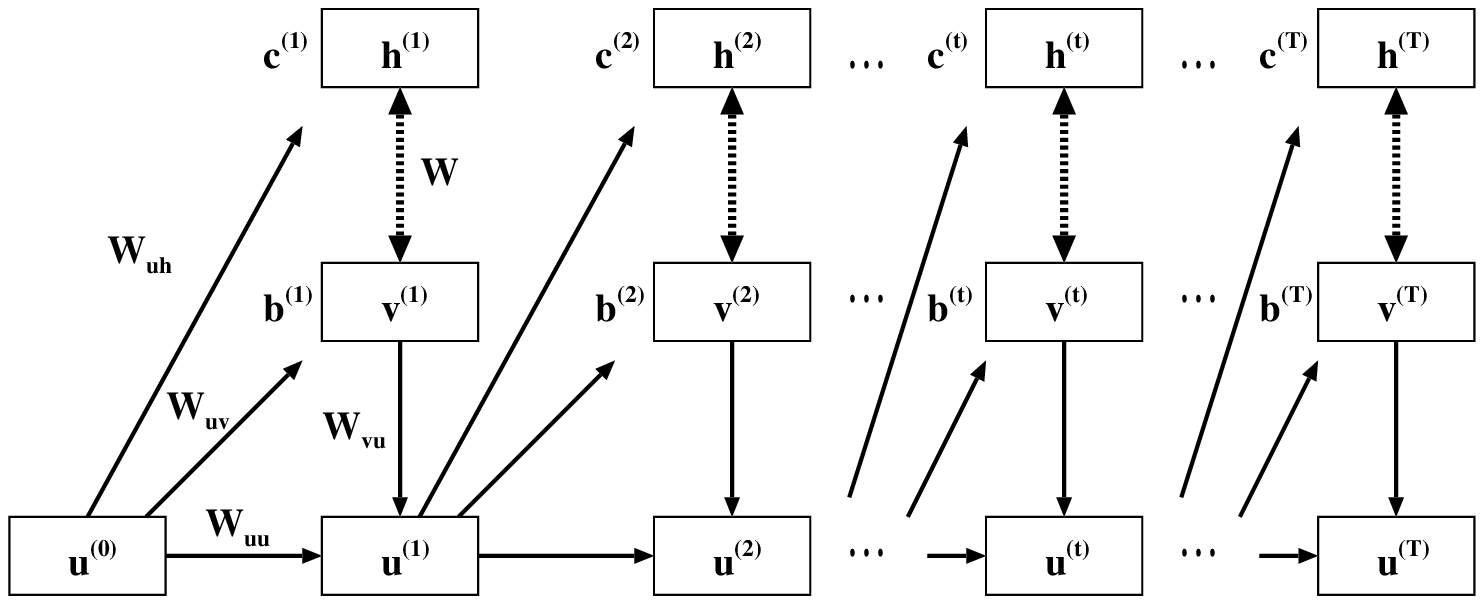}
\caption{Recurrent Neural Network RBM\cite{Lewandowski12}}
\label{fig:rnn-rbm}
\end{center}
\end{figure}

Recurrent Neural Networks RBM (RNN-RBM) \cite{Lewandowski12} as shown in Fig.~\ref{fig:rnn-rbm} is a similar model of RTRBM and realizes the specification of recurrent neural networks (RNNs). The network structure can incorporate an internal memory that can summarize the entire sequence history.

\begin{figure}[]
\begin{center}
\includegraphics[scale=0.6]{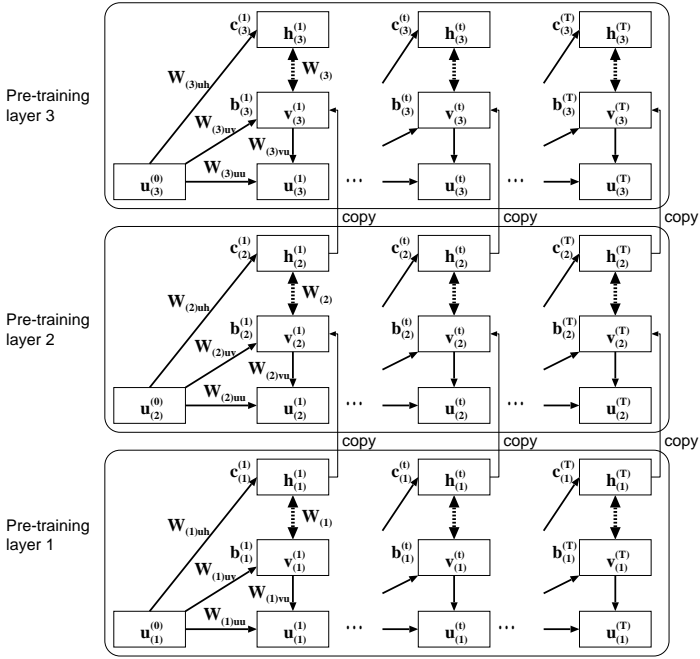}
\caption{Recurrent Neural Network DBN}
\label{fig:rnn-dbn}
\end{center}
\end{figure}

We propose the adaptive learning method in RNN-RBM to embed the neuron generation / annihilation described in subsection \ref{subsec:neurongeneration}. Moreover, we employ the hierarchical adaptive learning method of DBN that can determine the optimal number of hidden layers according to the given data set. The learning method can perform higher accuracy rate of classification to input patterns even if the input pattern has noise. Our proposed adaptive learning method of RNN-DBN as shown in Fig.~\ref{fig:rnn-dbn} has a deep architecture that can represent multiple features of input patterns hierarchically with the pre-trained adaptive learning method of RNN-RBM.


The RNN-RBM model has the state of representing contexts in time series data, $\bvec{u} \in \{0, 1 \}^{K}$, related to past sequences of time series data in addition to the visible neurons and the hidden neurons of the traditional RBM. Let the sequence of input sequence with the length $T$ be $\bvec{V} = \{ \bvec{v}^{(1)}, \cdots, \bvec{v}^{(t)}, \cdots, \bvec{v}^{(T)} \}$. The parameters $\bvec{b}^{(t)}$ and $\bvec{c}^{(t)}$ for the visible layer and the hidden layer, respectively are calculated from $\bvec{u}^{(t-1)}$ for time $t-1$ by using Eq.(\ref{eq:calc_bt}) and Eq.(\ref{eq:calc_ct}). The state $\bvec{u}^{(t)}$ at time $t$ is updated by using Eq.(\ref{eq:calc_ut}).

\begin{equation}
\label{eq:calc_bt}
\bvec{b}^{(t)} = \bvec{b} + \bvec{W}_{uv} \bvec{u}^{(t-1)}
\end{equation}
\begin{equation}
\label{eq:calc_ct}
\bvec{c}^{(t)} = \bvec{c} + \bvec{W}_{uh} \bvec{u}^{(t-1)}
\end{equation}
\begin{equation}
\label{eq:calc_ut}
\bvec{u}^{(t)} = \sigma( \bvec{u} + \bvec{W}_{uu} \bvec{u}^{(t-1)} + \bvec{W}_{vu} \bvec{v}^{(t)} ),
\end{equation}
where $\sigma()$ is a sigmoid function. Fig.\ref{fig:rnn-rbm} shows the flow of RNN-RBM. $\bvec{{u}^{(0)}}$ is the initial state which is given a random value. RNN-RBM is trained the weights between $\bvec{v}^{(t)}$ and $\bvec{h}^{(t)}$ by using $\bvec{b}^{(t)}$，$\bvec{c}^{(t)}$ at time $t$ and weights $\bvec{W}$. After the error are calculated till time $t$, the gradients for $\bvec{\theta}=\{\bvec{b}, \bvec{c}, \bvec{W}, \bvec{u}, \bvec{W}_{uv}, \bvec{W}_{uh}, \bvec{W}_{vu}, \bvec{W}_{uu} \}$ are updated to trace from time $T$ back to time $t$ by BPTT( Back Propagation Through Time) method. Algorithm \ref{alg:adaptive-rnn-rbm} shows the learning process of adaptive RNN-RBM.

In the similar way of the adaptive learning method of RBM, we monitor $\bvec{c}$ and $\bvec{W}$ in the learning. If Eq.(\ref{eq:neuron_generation}) and Eq.(\ref{eq:neuron_annihilation}) are satisfied, the neuron generation / annihilation algorithm works to find the optimal structure of RBM. Moreover, the structural learning mentioned in subsection \ref{sec:structurallearning} is applied to find the sparse structure.

The adaptive learning method of RNN-DBN is building the hierarchical network structure to pile up the pre-trained RNN-RBM in stages. In RNN-RBM, the hidden neuron $\bvec{h}^{(t)}$ for the input signal $\bvec{v}^{(t)}$ at time $t$ is calculated from $\bvec{u}^{(t-1)}$ at time $t-1$ deterministically. The output signals for the given input signals in RNN-RBM are determined uniquely and set them to the input signals in the subsequent layer in RNN-DBN are trained repeatedly. Algorithm \ref{alg:adaptive-rnn-dbn} shows the learning process of adaptive RNN-DBN.

\begin{algorithm}                      
\caption{Adaptive RNN-RBM}         
\label{alg:adaptive-rnn-rbm}                          
\begin{algorithmic}                  
\STATE Set initial parameter $\bvec{u}^{(0)}$.
\FORALL{$\bvec{v}^{(t)} (1 \leq t \leq T)$}
\STATE Calculate $\bvec{b}^{(t)}$ and $\bvec{c}^{(t)}$ from $\bvec{u}^{(t-1)}$ by Eq.(\ref{eq:calc_bt}) and Eq.(\ref{eq:calc_ct})).
\STATE Update $\bvec{u}^{(t)}$ from $\bvec{u}^{(t-1)}$ and $\bvec{v}^{(t)}$ by Eq.(\ref{eq:calc_ut})).
\ENDFOR

\STATE Calculate cost between $\bvec{v}^{(t)}$ and $\bvec{h}^{(t)}$ by CD-k.
\STATE Calculate gradients for the parameters $\bvec{\theta}$ by BPTT and update them.

\FORALL{hidden neuron $j$}
\IF{the neuron generation process is not completed}
\IF{the neuron generation condition is satisfied with Eq.(\ref{eq:neuron_generation})}
\STATE A new hidden neuron $j+1$ is generated and inserted.
\ENDIF
\ELSE{}
\IF{the neuron annihilation condition is satisfied with Eq.(\ref{eq:neuron_annihilation})}
\STATE The hidden neuron $j$ is annihilated.
\ENDIF
\ENDIF

\ENDFOR
\end{algorithmic}
\end{algorithm}

\begin{algorithm}                      
\caption{Adaptive RNN-DBN}
\label{alg:adaptive-rnn-dbn}                          
\begin{algorithmic}
\STATE Set $1$-th input $\bvec{V}_{(1)}$ and initial value of parameters $\bvec{\theta}_{(1)}$.
\FOR{$1 \leq l \leq L$}
\STATE Make pre-training $l$-th RBM for given $\bvec{V}_{(l)}$ and $\bvec{\theta}_{(l)}$.
\IF{the neuron layer conditions are satisfied with Eq.(\ref{eq:layer_generation1}) and Eq.(\ref{eq:layer_generation2}) during the learning}
\STATE The layer $l+1$ is generated.
\STATE Calculate $l$-th input $\bvec{V}_{(l+1)}$ and set initial value of parameters $\bvec{\theta}_{(l+1)}$.
\ELSE{}
\STATE The layer generation is stopped.
\ENDIF
\ENDFOR
  
\end{algorithmic}
\end{algorithm}

\section{Experimental Results for Benchmark Data Set}
\label{sec:EXE}

\begin{figure*}[]
\centering
\subfigure[Energy]{\includegraphics[scale=0.7]{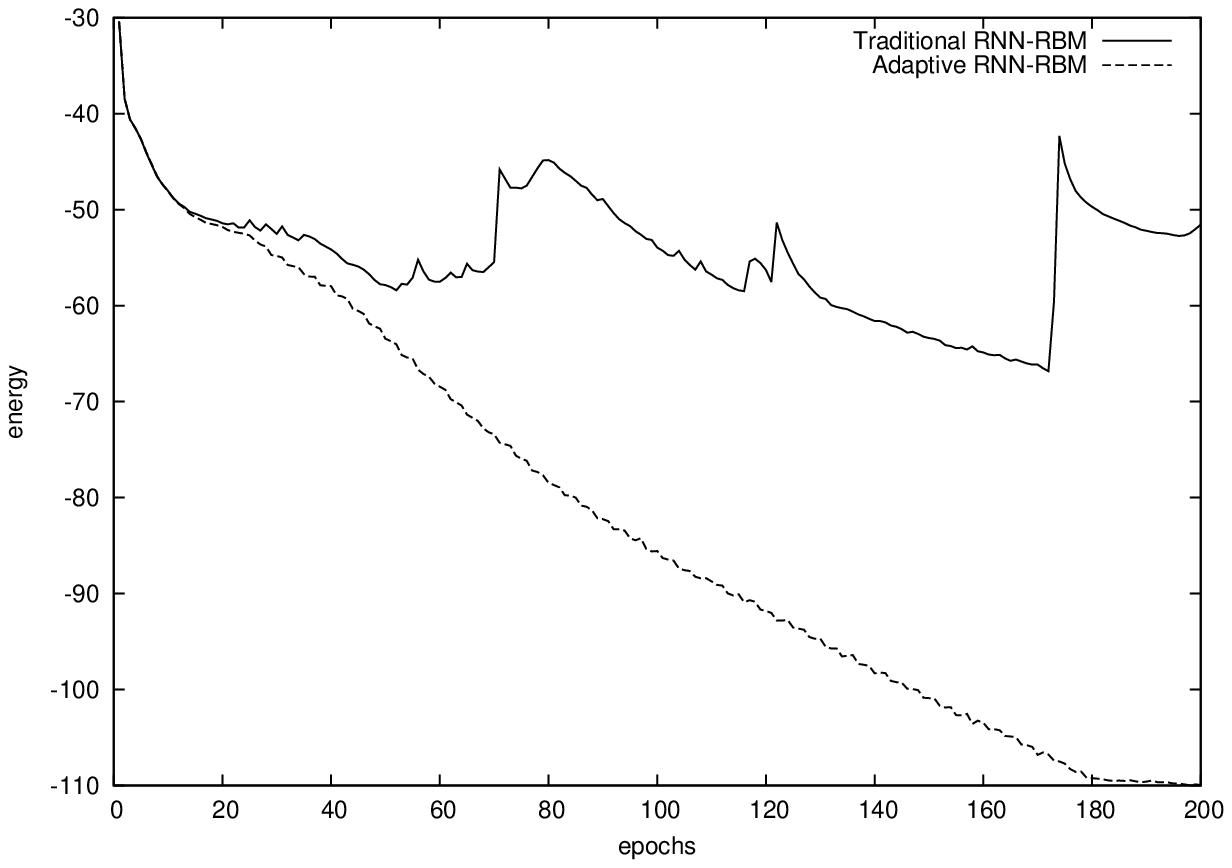}\label{fig:exe_layer1_energy_Nottingham}}
\subfigure[Error]{\includegraphics[scale=0.7]{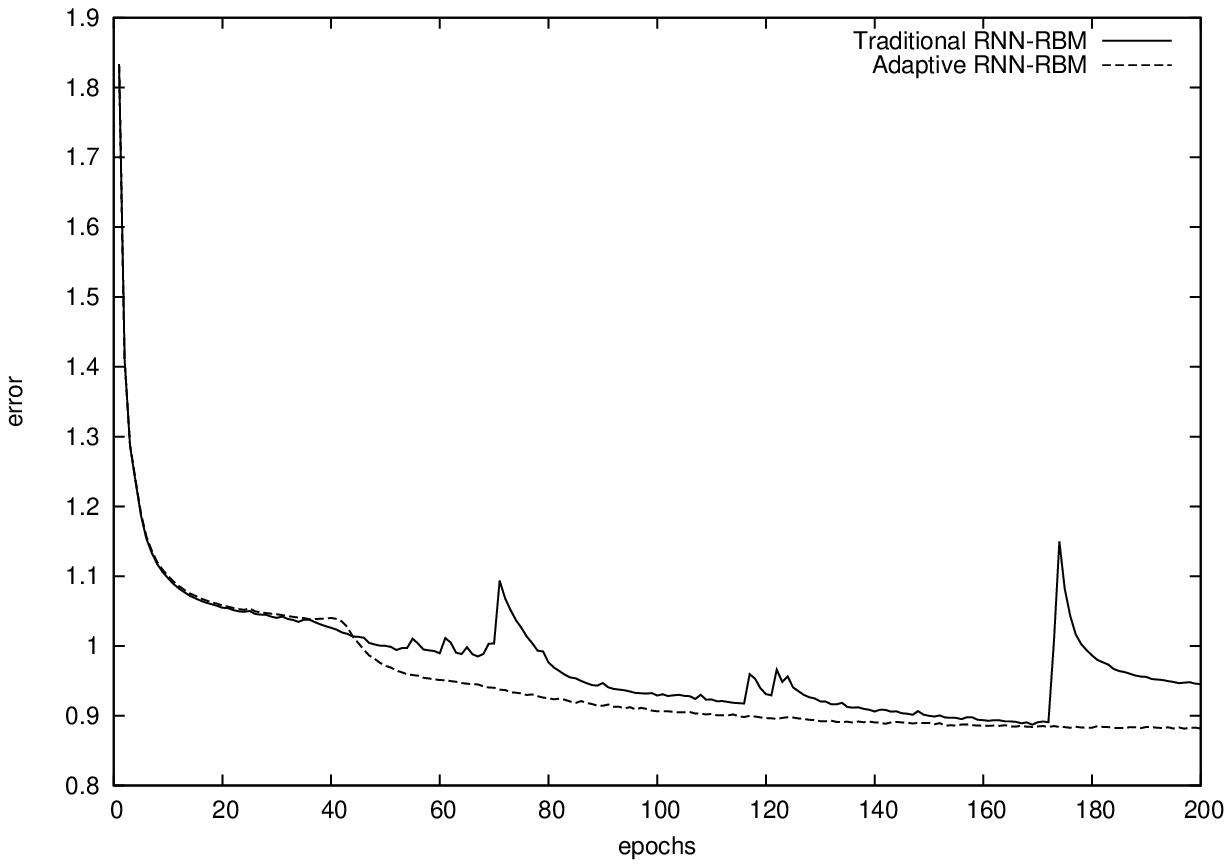}\label{fig:exe_layer1_error_Nottingham}}
\subfigure[Variance of $\bvec{W}$]{\includegraphics[scale=0.7]{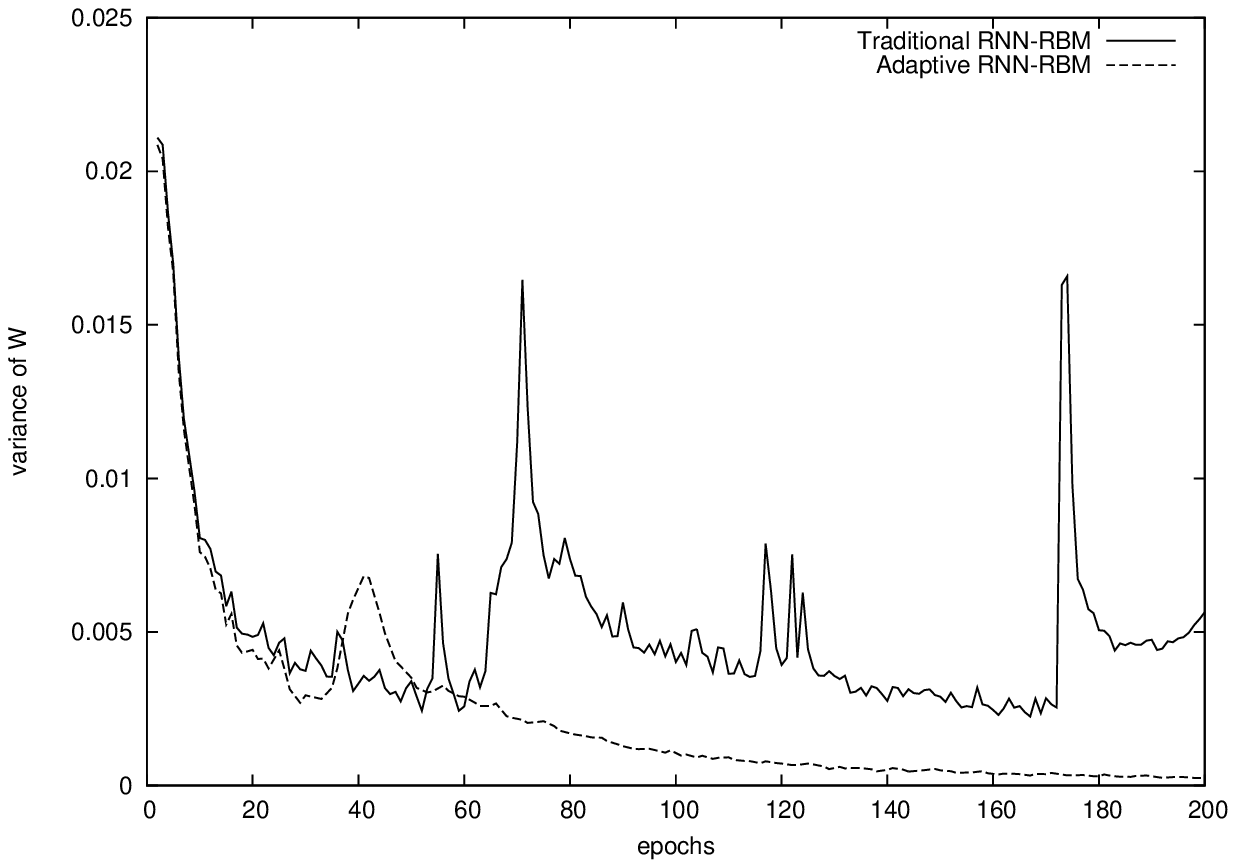}\label{fig:exe_layer1_W_Nottingham}}
\subfigure[Variance of $\bvec{c}$]{\includegraphics[scale=0.7]{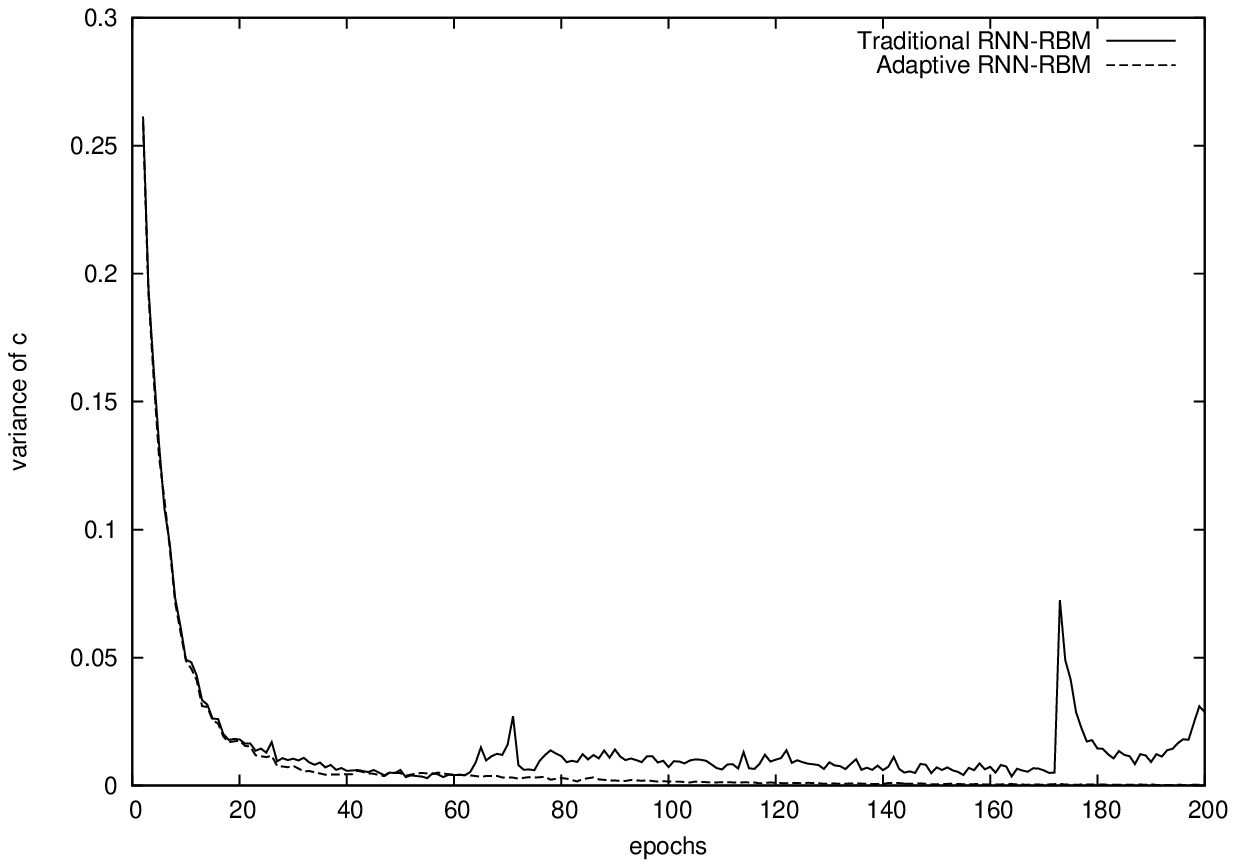}\label{fig:exe_layer1_c_Nottingham}}
\caption{Experimental Results on Nottingham (layer 1)}
\label{fig:exe_layer1_Nottingham}
\end{figure*}

\begin{table*}[htb]
\caption{Prediction Accuracy on Nottingham}
\vspace{-5mm}
\label{tab:accuracy_nottingham}
\begin{center}
\begin{tabular}{l|r|r|r|r}
\hline \hline
& No. layers & Error(Training) & Error(Test) & Correct ratio (Test) \\ \hline\hline
Traditional RNN-RBM              & 1   &  0.945  &  1.704  &  71.7\% \\ \hline
Adaptive  RNN-RBM                & 1   &  {\bf 0.881}  &  {\bf 1.240}  &  {\bf 76.5}\% \\ \hline \hline
Traditional RNN-DBN              & 4   &  0.217    & 1.381 &  75.8\% \\ \hline 
Adaptive  RNN-DBN                & 4   &  {\bf 0.101}    &  {\bf 0.133} &  {\bf 85.3}\% \\ \hline
\hline 
\end{tabular}
\end{center}
\end{table*}

\begin{figure*}[t]
\centering
\subfigure[Energy]{\includegraphics[scale=0.7]{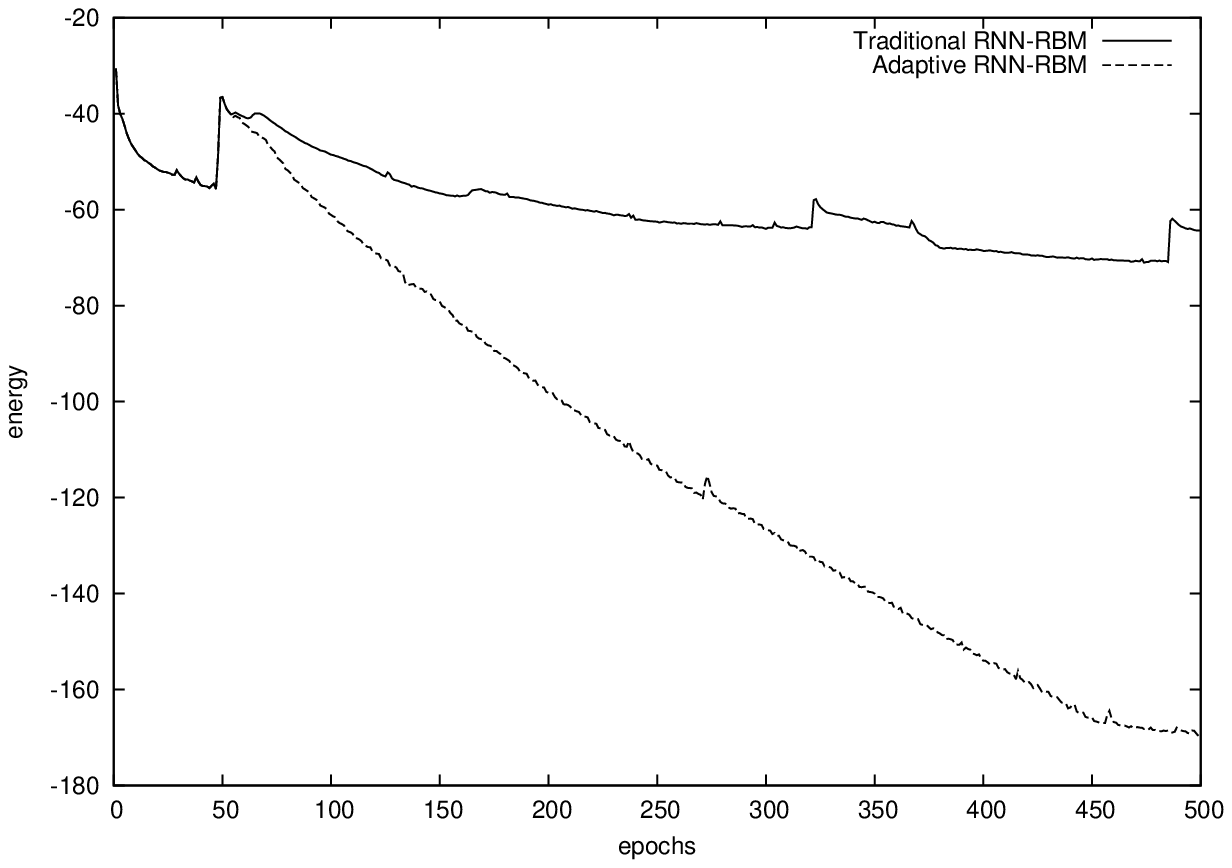}\label{fig:exe_layer1_energy_CMU}}
\subfigure[Error]{\includegraphics[scale=0.7]{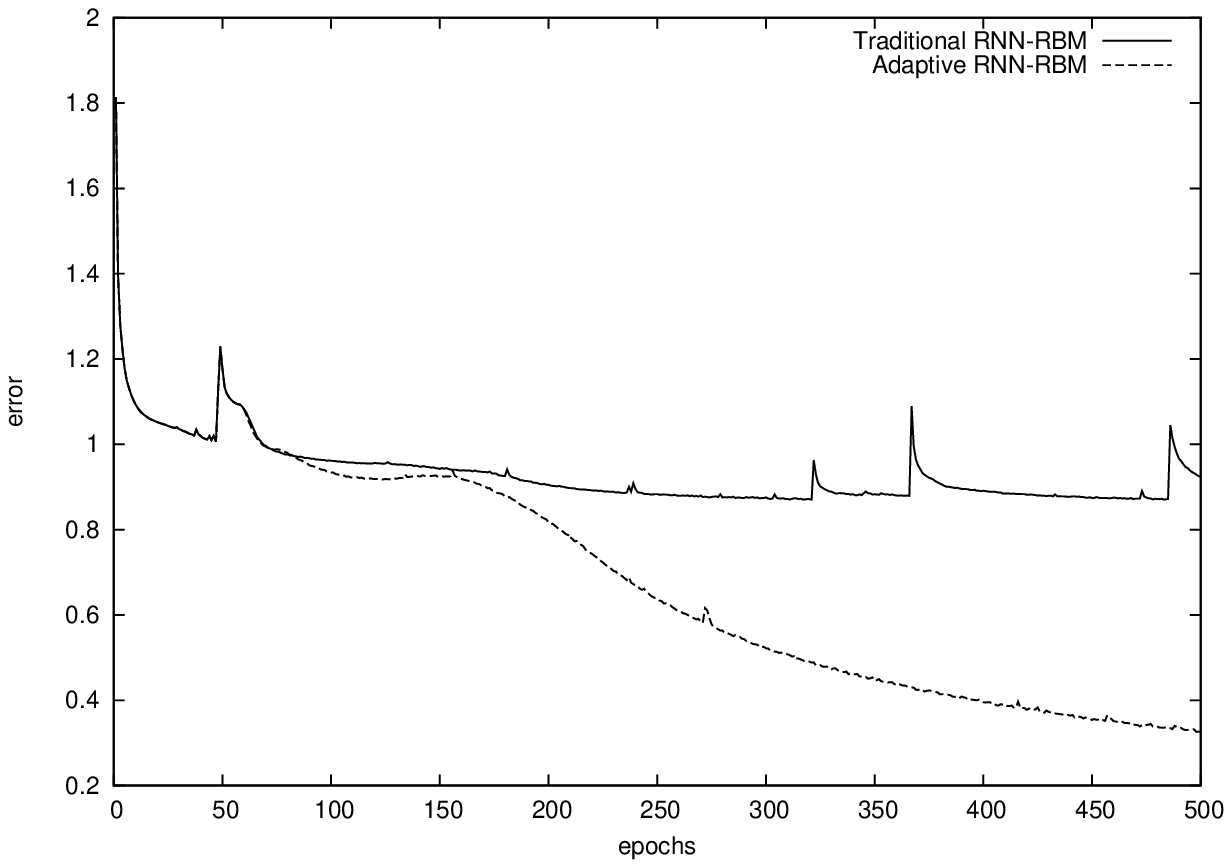}\label{fig:exe_layer1_error_CMU}}
\subfigure[Variance of $\bvec{W}$]{\includegraphics[scale=0.7]{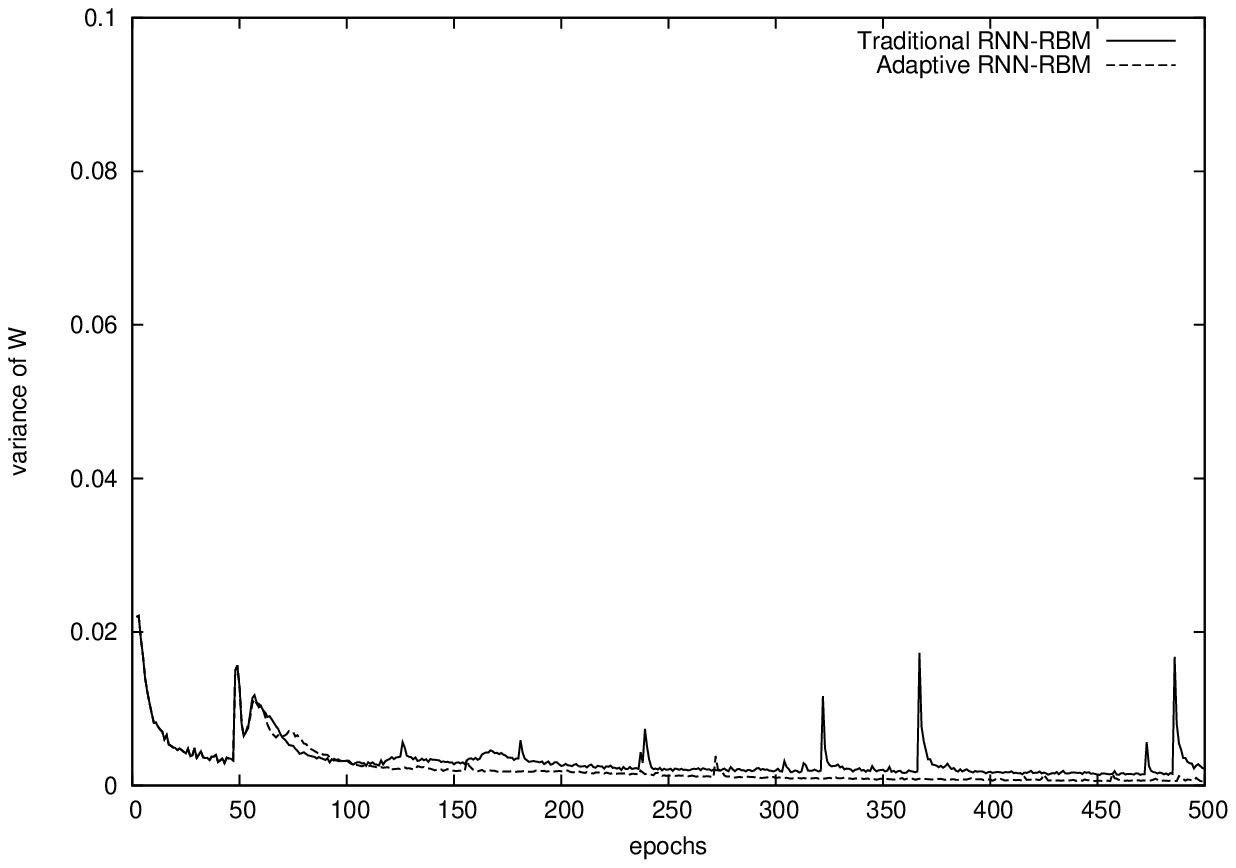}\label{fig:exe_layer1_W_CMU}}
\subfigure[Variance of $\bvec{c}$]{\includegraphics[scale=0.7]{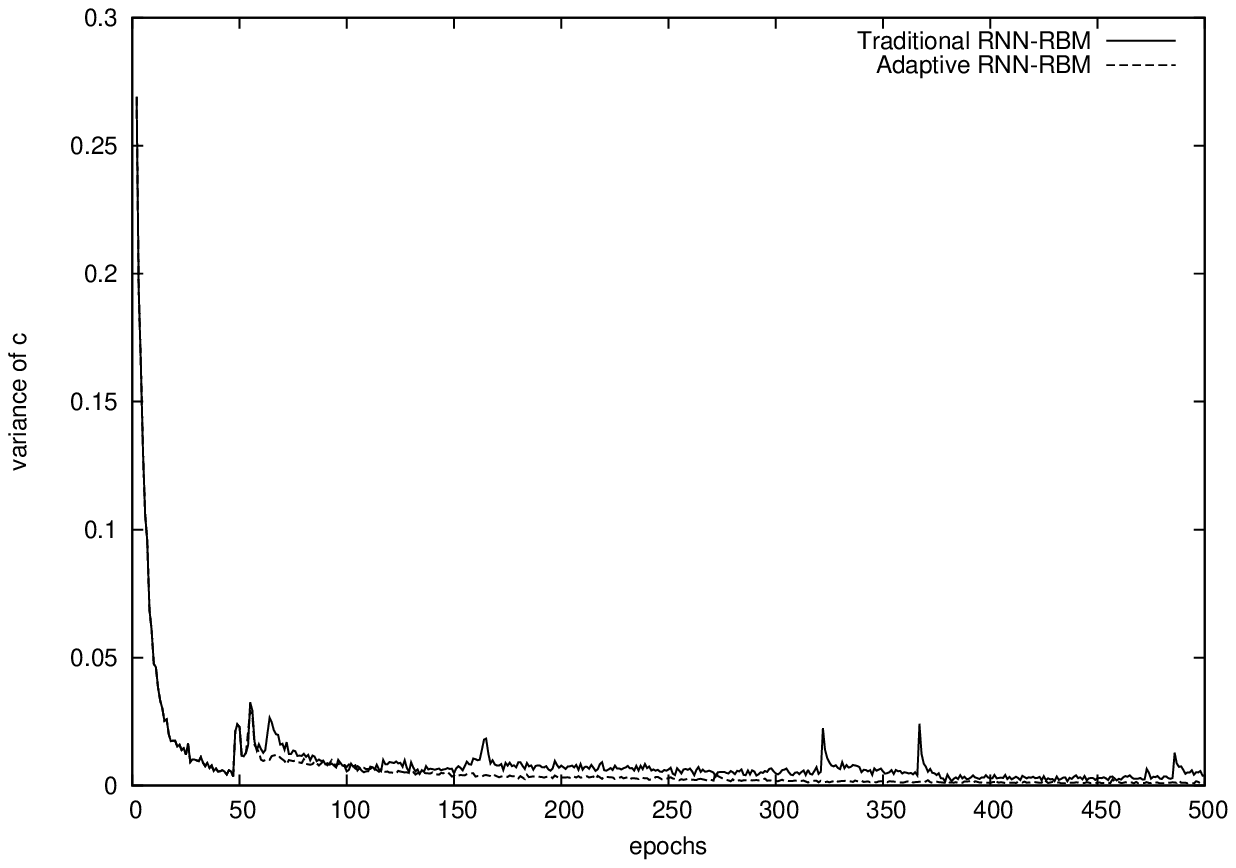}\label{fig:exe_layer1_c_CMU}}
\caption{Experimental Results on CMU (layer 1)}
\label{fig:exe_layer1_CMU}
\end{figure*}

\begin{table*}[htb]
\caption{Prediction Accuracy on CMU}
\vspace{-5mm}
\label{tab:accuracy_cmu}
\begin{center}
\begin{tabular}{l|r|r|r|r}
\hline \hline
& No. layers & Error(Training) & Error(Test) & Correct ratio (Test) \\ \hline\hline
Traditional RNN-RBM              & 1   &  1.344  &  2.401  &  65.2\% \\ \hline
Adaptive  RNN-RBM                & 1   &  {\bf 0.981}  &  {\bf 1.471}  &  {\bf 73.1}\% \\ \hline \hline
Traditional RNN-DBN              & 6   &  0.517    & 2.202 &  70.8\% \\ \hline 
Adaptive  RNN-DBN                & 6   &  {\bf 0.121}    &  {\bf 0.148} &  {\bf 82.3}\% \\ \hline
\hline 
\end{tabular}
\end{center}
\end{table*}

\subsection{Data Sets}
In this experiments, the benchmark data set `Nottingham' \cite{Nottingham} and `CMU' \cite{CMU} were used to verify the effectiveness of our proposed method. Nottingham is a classical piano MIDI archive included in about 694 training cases and about 170 test cases. On the other hand, CMU is a motion capture database collected by Carnegie Mellon University. There are 2,605 trials in 6 categories which are divided into 23 subcategories. We use the same Benchmark test used in \cite{Sutskever08} and \cite{Lewandowski12} to compare the capability of our proposed method with their methods.

The parameters used in this paper are as follows. The training algorithm is Stochastic Gradient Descent (SGD) method, the batch size is 100, the learning rate is 0.01, $\theta_{G} = 0.001$, $\theta_{A} = 0.1$, $\theta_{L1} = 0.01$, $\theta_{L2} = 0.01$, $\epsilon_{1} = \epsilon_{2} = \epsilon_{3} = 0.001$, $\theta = 0.1$. We used the computer with the following specifications: CPU = Intel(R) 24 Core Xeon E5-2670 v3 2.3GHz, GPU = Tesla K80 4992 24GB times 3, Memory = 64GB, OS = Centos 6.7 64 bit.

\subsection{Experimental Results}
Fig.~\ref{fig:exe_layer1_Nottingham} shows the learning situation of the RNN-RBM \cite{Lewandowski12} and the adaptive RNN-RBM on Nottingham. Fig.~\ref{fig:exe_layer1_energy_Nottingham} - Fig.~\ref{fig:exe_layer1_c_Nottingham} are the learning curve of energy function and error function, the variance of $\bvec{W}$, the variance of $\bvec{c}$, and the change of the number of hidden neurons. As shown in Fig.~\ref{fig:exe_layer1_W_Nottingham} and Fig.~\ref{fig:exe_layer1_c_Nottingham}, the variance of the parameters was significantly fluctuated even after a certain period of the learning in the RNN-RBM. As a result, the energy function and the error function were not converged with smaller value as shown in Fig.~\ref{fig:exe_layer1_energy_Nottingham} and Fig.~\ref{fig:exe_layer1_error_Nottingham}. On the other hand, the neuron generation process in our proposed RNN-RBM was operated after 20th iterations. After the generation process, 52 additional hidden neurons were generated (total number is 62) and the network structure enough to the input patterns was obtained. As a result, the energy function and error function were converged into a small value. After the learning of RNN-RBM, a structural learning method with forgetting was applied to get the sparse structure of RBM with 42 hidden neurons. The RNN-DBN builds the hierarchical network structure to pile up the trained RNN-RBMs. The total CPU time to the end of computation was 529 and 413 minutes for the traditional RNN-DBN and the adaptive RNN-DBN, respectively.

Such an observation can be also seen on the training result on CMU as shown in Fig.~\ref{fig:exe_layer1_CMU}. In the case of CMU, there are 100 hidden neurons in the beginning of learning and 100 additional hidden neurons were generated during the learning, then the total number of hidden neuron was 200. After a structural learning method with forgetting, there were 167 hidden neurons. The total CPU time to the end of computation was 151 and 125 minutes for the traditional RNN-DBN and the adaptive RNN-DBN, respectively.

Table~\ref{tab:accuracy_nottingham} and Table~\ref{tab:accuracy_cmu} show the prediction accuracy on Nottingham and CMU, respectively. Our proposed adaptive RNN-RBN and RNN-DBN can obtain a higher prediction accuracy for not only training set but also test set than traditional RNN-RBM and RNN-DBN model. The prediction accuracy rate are 76.4\% for Nottingham test set, 73.2\% for CMU test set, which the value are traditional DBN model. It is remarkable that adaptive learning method of RNN-RBM performs better than the traditional RNN-DBN, because the pre-training at lower layer in the traditional RNN-DBN cannot work well to represent abstract features for input patterns. The problem will be solved by finding the optimal setting of parameters, however, a try and error operation should be required. As a result, the adaptive RNN-DBN can obtain the highest prediction accuracy with smallest energy and error for test set.

\section{Conclusive Discussion}
The adaptive structural learning method of RBM is to self-organize the optimal network structure in terms of energy stability as well as clarification of knowledge according to the given input data during the learning phase. Moreover, we developed assemble method of pre-trained RBM by using the layer generation condition in hierarchical DBN. The proposed method is superior at the self-organized method to train the optimal network structure for given training data set during training process. Our proposed adaptive learning method of DBN records a great score as for image classification task. However, the previous method showed the classification capability of static image data enough, but the classification for the time-series data was not high. In this paper, we embed the adaptive learning method into the RNN-RBM model. Moreover, RNN-RBM is extended to realize the DBN model with self-generated layer algorithm and then the model shows the high prediction accuracy for the time-series data sets. We will apply our proposed method into another time-series data set such as medical image data such mammography data in future works.

\section*{Acknowledgment}
This work was supported by JAPAN MIC SCOPE Grand Number 162308002.

\end{document}